%% file: main.tex
\pdfoutput=1

\PassOptionsToPackage{dvipsnames}{xcolor}

\documentclass[11pt]{article}
\usepackage[]{./template/naacl2021/naacl2021}

\usepackage{times}
\usepackage{latexsym}

\usepackage[T1]{fontenc}
\usepackage[utf8]{inputenc}

\PassOptionsToPackage{numbers}{natbib}
\usepackage{graphicx}
\graphicspath{{./figure/}}
\usepackage{bm}
\usepackage{amsmath}
\usepackage{multirow}
\usepackage{placeins}
\usepackage{subcaption}
\usepackage{comment}

\usepackage{soul}
\usepackage{amssymb}
\usepackage{array}
\input{./template/math_commands}

\usepackage{microtype}
\usepackage{booktabs} 

\usepackage{hyperref}

\usepackage[noend]{algorithmic}
\usepackage{./template/icml2020/algorithm}
\renewcommand\algorithmicthen{}
\renewcommand\algorithmicdo{}

\title{Self-Supervised Contrastive Learning for \\Efficient User Satisfaction Prediction in Conversational Agents}

\author{Mohammad Kachuee\thanks{~~Work done as an intern at Amazon Alexa AI.} \\
  UCLA Computer Science \\
  Los Angeles, CA \\
  \texttt{mkachuee@cs.ucla.edu} \\
  \And
  Hao Yuan \\
  Amazon Alexa AI \\
  Seattle, WA \\
  \And
  Young-Bum Kim \\
  Amazon Alexa AI \\
  Seattle, WA \\
  \texttt{\{yuanha,youngbum,sungjinl\}@amazon.com} \\  
  \And
  Sungjin Lee\\
  Amazon Alexa AI \\
  Seattle, WA \\
  }

\date{}

\begin{document}

\maketitle

\begin{abstract}
Turn-level user satisfaction is one of the most important performance metrics for conversational agents. It can be used to monitor the agent's performance and provide insights about defective user experiences. 
While end-to-end deep learning has shown promising results, having access to a large number of reliable annotated samples required by these methods remains challenging. In a large-scale conversational system, there is a growing number of newly developed skills, making the traditional data collection, annotation, and modeling process impractical due to the required annotation costs and the turnaround times.
In this paper, we suggest a self-supervised contrastive learning approach that leverages the pool of unlabeled data to learn user-agent interactions. We show that the pre-trained models using the self-supervised objective are transferable to the user satisfaction prediction. In addition, we propose a novel few-shot transfer learning approach that ensures better transferability for very small sample sizes. The suggested few-shot method does not require any inner loop optimization process and is scalable to very large datasets and complex models.
Based on our experiments using real data from a large-scale commercial system, the suggested approach is able to significantly reduce the required number of annotations, while improving the generalization on unseen skills.
\end{abstract}

\section{Introduction}
Nowadays automated conversational agents such as Alexa, Siri, Google Assistant, Cortana, etc. are widespread and play an important role in many different aspects of our lives. Their applications vary from storytelling and education for children to assisting the elderly and disabled with their daily activities. Any successful conversational agent should be able to communicate in different languages and accents, understand the conversation context, analyze the query paraphrases, and route the requests to various skills available for handling the user's request \citep{ram2018conversational}.

In such a large-scale system with many components, it is crucial to understand if the human user is satisfied with the automated agent's response and actions. In other words, it is desirable to know if the agent is communicating properly and providing the service that is expected by the user. In the literature, it is referred to as targeted turn-level satisfaction as we are only interested in the user's satisfaction for a certain conversation turn given the context of the conversation, and not the overall satisfaction for the whole conversation \citep{park2020large}. Perhaps the most basic use of a user satisfaction model would be to monitor the performance of an agent and to detect defects as a first step to fix issues and improve the system. Anticipating user dissatisfaction for a certain turn in a conversation, an agent would be able to ask the user for repeating the request or providing more information, improving the final experience. Also, a powerful user satisfaction model can be used as a ranking or scoring measure to select the most satisfying response among a set of candidates and hence guiding the conversation.

The problem of user satisfaction modeling has recently attracted significant research attention \citep{jiang2015automatic,bodigutla2019multi,park2020large,pragst2017recurrent,rach2017interaction}. These methods either rely on annotated datasets providing ground-truth labels to train and evaluate \citep{bodigutla2019multi} or rely on ad hoc or human-engineered metrics that do not necessarily model the true user satisfaction \citep{jiang2015automatic}. Access to reliable annotations to be used in building satisfaction models has been very challenging partly due to the fact that a large-scale conversation system supports many different devices as well as voice, language, and application components, providing access to a wide variety of skills. The traditional approach of collecting samples from the live system traffic and tasking human annotators to label samples would not be scalable due to the cost of annotations as well as the turn-around time required to collect and annotate data for a new skill or feature. Note that onboarding new skills in a timely manner is a crucial to ensure active skill developer engagement.

To address this problem, we propose a novel training objective and transfer learning scheme that significantly improves not only the data efficiency but also the model generalization to unseen skills. In summary, we make the following contributions:
\begin{itemize}
    \item We propose a contrastive self-supervised training objective that can leverage virtually any unlabeled conversation data to learn user-agent interactions.
    \item We show that the proposed method can be used to pre-train state-of-the-art deep language models and the acquired knowledge is transferable to the user satisfaction prediction.
    \item We suggest a novel and scalable few-shot transfer learning approach that is able to improve the label efficiency even further in the case of few-shot transfer learning. 
    \item We conduct extensive experiments using data from a large-scale commercial conversational system, demonstrating significant improvements to label efficiency and generalization.
\end{itemize}

\section{Related Work}
\label{sec:Related Work}
\subsection*{User Satisfaction in Conversational Systems}

The traditional approach to evaluating a conversational system is to evaluate different functionalities or skills individually. For instance, for a knowledge question answering or web search skill, one can use response quality metrics commonly used to evaluate search system and ranking systems such as nDCG \citep{jarvelin2008discounted,hassan2012semi,fox2005evaluating}. While these methods provide justifiable measures for certain skills, they are not extendable to a large number of skills, especially for skills without a set of proper hand-engineered features and metrics, or newly developed third-party skills \citep{bodigutla2019multi}.

Another, more general, line of research is to evaluate the performance of a conversation system from the language point of view. Here, the objective is to measure how natural, syntactically and semantically, an automated agent is able to interact with a human user. For instance, using generic metrics such as BLEU~\citep{papineni2002bleu} or ROUGE~\citep{lin2004rouge} one can measure how the agent's responses are consistent with a set of provided ground-truth answers. However, these approaches not only suffer from shortcomings such as inconsistency with the human understanding \citep{liu2016not,novikova2017we} but also are not practical for a real-world conversation system due to their dependence on ground-truth responses.

A more recent approach is to use human annotations specifically tailored for the user satisfaction task as a source of supervision to train end-to-end prediction models~\citep{bodigutla2019multi}. \citet{jiang2015automatic} suggested training individual models for 6 general skills and devised engineered features to link user actions to the user satisfaction for each studied skill. \citet{park2020large} proposed a hybrid method to learn from human annotation and user feedback data that is scalable and able to model user satisfaction across a large number of skills.

\subsection*{Contrastive Learning}
\citet{gutmann2010noise} was the first study to propose the idea of noise-contrastive learning in the context of a capturing a distribution using an objective function to distinguish samples of the target distribution from samples of an artificially generated noise distribution. Contrastive predictive coding (CPC) \citep{oord2018representation} suggested the idea of using an NCE objective to train an auto-regressive sequence representation model. Deep InfoMax~\citep{hjelm2018learning} used self-supervised contrastive learning in an architecture where a discriminator is trained to distinguish between representations of the same image (positive samples) or representations of different images (negative samples). While many different variations of contrastive methods have been suggested, the main idea remains the same: defining a self-supervised objective to distinguish between the hidden representations of samples from the original distribution and samples from a noise distribution \citep{trinh2019selfie,devon2020representation,yao2020self}.

\subsection*{Few-shot Transfer Learning}

Few-shot transfer learning is a very active and broad subject of research. We limit the scope of our study to methods in which a form of gradient supervision is provided by a target task to ensure the efficient transferability of representations trained on a source task.
\citet{lopez2017gradient} suggested the idea of joint multi-task training and using the cosine similarity of the concatenated network gradients from the source and target tasks. For gradients with negative cosine distance, they project the source gradients to a more aligned direction by solving a quadratic programming problem. \citet{luo2020n} continued that line and suggested a method in the context of few-shot transfer learning, showing that using even a few samples from the target task can significantly improve the transferability of the trained models. \citet{li2020gradmix} presented a similar idea but suggested adjusting learning rates for each layer to improve the cosine similarity of different tasks. While these methods show promising results, they only measure the similarity between concatenated gradient vectors consisting of all network parameters which is a very rough measure of alignment. Also, they require solving for a quadratic or iterative optimization problem as an inner loop in the training procedure that can be computationally expensive and often prohibitive for large-scale problems.

\section{User Satisfaction Modeling}

\begin{figure*}[t]
    \centering
        \includegraphics[width=0.8\linewidth]{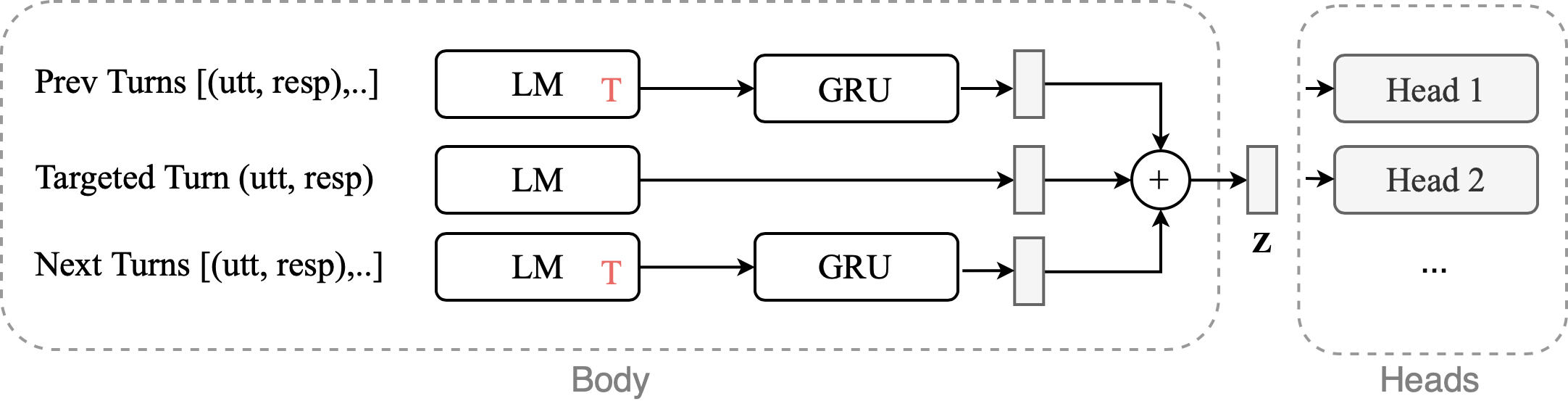}
        \caption{Overview of the suggested network architecture. In our architecture, BERT encoder with average pool at the last layer is used as the LM. We consider a context window of at most 2T+1 turns. Heads are simple MLPs classifiers with one hidden layer.}
        \label{fig:arch}
\end{figure*}

\subsection{Problem Definition}

In this paper, we consider the conversational interaction between a human user and an automated agent. Each interaction consists of a set of turns in which the user provides an utterance and the agent provides appropriate responses. A set of turns that are happening within a certain time window are grouped as a conversation session. Formally, we can represent a session as a set of turns:
\begin{equation}
    S_i = \{ (U_i^{t=0},R_i^{t=0}), \dots, (U_i^{t=T},R_i^{t=T}) \}
\end{equation}
Here, $S_i$ represents session $i$ consisting of a set of turns as tuples of utterance and responses, $(U_i^t,R_i^t)$, for the first turn $t=0$ to the last turn $t=T$ in that session.

In the context of turn-level user satisfaction modeling, we are interested in the classification of a certain targeted turn within a session as either satisfying (SAT) or dissatisfying (DSAT). Note that the satisfaction here is defined based on the agent's response given a certain utterance and the context (i.e., other session turns). We use the notation $Y_i^{t^{*}} \in \{\text{SAT},\text{DSAT}\}$ to indicate the user satisfaction for the targeted turn $t=t^{*}$ of session $i$. See Figure~\ref{fig:sat_example} for examples of SAT/DSAT interactions.

\begin{figure}[H]
    \centering
    \begin{subfigure}[b]{0.99\columnwidth}
        \centering
        \resizebox{\columnwidth}{!}{
        \begin{tabular}{ll}
        \hline
        \textbf{Session} & \textbf{Label} \\
        \hline
        \textbf{U0:} \textit{play clocks} \\
        \textbf{R0:} \textit{Playing clocks by Coldplay.} & \textcolor{ForestGreen}{\textbf{SAT}}\\
        \textit{{[}User listening for 1 minute]} & \small{(user listening)}\\
        
        \hline
        \textbf{U0:} \textit{play clock} \\
        \textbf{R0:} \textit{Sorry, I cannot find the song.} &  \textcolor{BrickRed}{\textbf{DSAT}}\\
        {U1:} \textit{play clocks} & \small{(repeated request)}\\
        {R1:} \textit{Playing clocks by Coldplay.} & \\
        \textit{{[}User listening for 1 minute]} \\
        
        \hline
        \textbf{U0:} \textit{turn on the tv} \\
        \textbf{R0:} \textit{Did you mean Living Room TV?} &  \textcolor{ForestGreen}{\textbf{SAT}}\\
        {U1:} \textit{yes} \\
        {R1:} \textit{[EMPTY]} & \small{(no further request)}\\

        \hline
        \textbf{U0:} \textit{call chris} \\
        \textbf{R0:} \textit{Calling chris.} &  \textcolor{BrickRed}{\textbf{DSAT}}\\
        {U1:} \textit{stop} & \small{(ASR error, barge-in)}\\
        {R1:} \textit{[EMPTY]} & \\        
        {U1:} \textit{call christina} \\
        {R1:} \textit{Calling Christina.} & \\
        \textit{{[}Phone call for 2 minutes]} \\
        \bottomrule
        
        \end{tabular}
        }
    \end{subfigure}
    \caption{A few examples of \textcolor{ForestGreen}{SAT} and \textcolor{BrickRed}{DSAT} turns to illustrate the importance of the conversation context.}
    \label{fig:sat_example}
    \vspace{-0.0in}
\end{figure}

\subsection{Datasets\protect\footnote{Due to confidentiality concerns, we are not able to disclose the exact annotation protocols and data specifications.}}

In this study, we use real-world data from Alexa, a large-scale commercial conversational agent. Specifically, we use a dataset of about 891,000 real-world conversation sessions in which a certain turn within each session is annotated by a human annotator as SAT or DSAT. Human annotators had access to the session context and followed a standard labeling protocol (further information is provided in Appendix~\ref{sec:appendix_annotation}). As a preprocessing step, we limited turns within each session to a window of five turns: at most two turns before the targeted turn, the targeted turn, and at most two turns after the targeted turn. This labeled dataset is denoted as $\mathbb{D}_{sup}$.

In addition to $\mathbb{D}_{sup}$, we also use a large pool of real-world session data without any annotation or label. This dataset is about twice the size of $\mathbb{D}_{sup}$, but as we are not limited to targeted turns, we keep all session turns and decide context windows based on a randomized data augmentation step. The resulting effective sample size is significantly larger than $\mathbb{D}_{sup}$. We denote this unlabeled dataset as $\mathbb{D}_{unsup}$. As both datasets were sampled from real traffic, we ensured that there is no overlap between $\mathbb{D}_{unsup}$ and the evaluation splits of $\mathbb{D}_{sup}$.

The conversations cover a wide variety of internally developed (1p) and third-party (3p) developer skills. Due to the imbalanced traffic, in our datasets, there is a huge variation between the number of samples for different skills. For instance, 1p skills such as music or weather have hundreds of thousands of samples while many 3p skills only have less than 10 samples throughout our datasets. To properly evaluate the performance of our predictors on such imbalanced data, we proposed a novel approach to split the data and to evaluate. We build two test sets: a test set measuring in-domain performance and another test set to measure the out-of-domain generalization. The in-domain test set consists of samples from skills that the train set covers. The out-of-domain test set measures the performance on skills that are not covered by the train set. Ideally, we would like to observe good classification performance in both test splits, indicating the ability of our models to learn and model the current major traffic and to generalize to less frequent or future traffic. Based on this, we split $\mathbb{D}_{sup}$ to 70\% train, 15\% validation, and the rest for the test (about $1/5$ of test samples are out-of-domain and $4/5$ are in-domain). The in-domain and out-of-domain test sets consist of 17 and 275 skills, respectively. The $\mathbb{D}_{unsup}$ is randomly split to 80\% train and the rest for validation, regardless of skills. Table~\ref{tab:dataset_stats} presents a summary of dataset statistics for $\mathbb{D}_{sup}$.

\begin{table}[t]
    \centering
        \begin{tabular}{ll}
        \hline
        \textbf{Property} & \textbf{Size} \\
        \hline
        Total number of samples & $\approx 891,000$ \\
        Total number of 1p skills & $> 20$ \\
        Total number of 3p skills & $> 1500$ \\
        Ratio of SAT to DSAT samples & $> 20$ \\
        \bottomrule
        \end{tabular}
    \caption{Dataset statistics for $\mathbb{D}_{sup}$}
    \label{tab:dataset_stats}
\end{table}


\subsection{Network Architecture}
\label{sec:arch}

Figure~\ref{fig:arch} shows a high-level drawing of the network architecture used in our experiments. It consists of a language model (LM) that encodes utterance and response pairs to vector representations. Here, we consider up to T turns before and after the targeted turn. To further summarize the list of the previous or next turns, we use GRU layers \citep{chung2014empirical}. Then, an average pool is used to produce a representation vector, $\bm{z}$, for each session. Note that before the pooling, simple non-linear MLPs are used to transform each partial representation. Finally, $\bm{z}$ is used as an input to a set of different head networks, responsible for making predictions for different objectives.

Regarding the LM, we use the standard BERT encoder \citep{devlin2018bert} architecture pre-trained as suggested by \citet{liu2019roberta}. To make a fixed-length representation of the utterance response pairs i.e. turn semantics, we use an average pool at the last encoder layer of the BERT token representations. We also tried other approaches such as using the classification token instead of pooling, but based on our initial results simple pooling performed consistently better.

We share our BERT-based LM parameters across the network to encode the session turns. However, we train separate GRU networks to summarize the previous and next turns. The output dimension of the LM is equal to $768$, the size of the standard BERT hidden layer. The hidden layer and output size of our GRUs are $256$, and we use 2-layer bi-directional GRUs. Each head is a simple MLP with a single hidden layer of size $256$ followed by a ReLU nonlinearity. The final network consists of about $117.7$ million parameters from which about $110$ million is related to BERT and the rest is for GRUs, heads, etc.

\subsection{Supervised Learning Baseline}
\label{sec:sup}
As a baseline approach, we use the network defined in Section~\ref{sec:arch} with a binary classification head to distinguish SAT and DSAT samples. Here, we use labels provided by $\mathbb{D}_{sup}$ and a binary cross-entropy (BCE) loss function. An Adam optimizer \citep{kingma2014adam} with a batch size of $512$ is used to train the network for $10$ epochs. The base learning rate for all non-BERT layers is set to $10^{-3}$, while for BERT layers, we use a smaller learning rate of $5 \times 10^{-5}$. The learning rates are decayed with a factor $5$ twice at $60\%$ and $80\%$ of total iterations. Unless indicated otherwise, we use a similar training setup for other experiments suggested in this paper.

\section{Self-Supervised Contrastive Learning}

\subsection{Self-Supervised Objective}
\label{sec:nce}
We define a self-supervised objective in which the model is tasked to distinguish real sessions from unreal (or noisy) sessions. Any unlabeled dataset, such as $\mathbb{D}_{unsup}$ can be used to sample real sessions. To generate unreal textual information, different approaches have been suggested in the literature such as back-translation~\citep{fang2020cert}, generative modeling~\citep{liu2020self}, or even random word substitutions. 

In this work, we leverage the multi-turn and structured nature of sessions to generate noise samples by simply shuffling the targeted utterances/responses within each training batch (see Figure~\ref{fig:nce_example} for an example). Intuitively, the noise samples are sessions in which the targeted utterance or response does not belong to the rest of the session. Therefore, the model has to capture the joint distribution of the context and targeted turns.
Algorithm~\ref{alg:nce} shows an overview of the sample generation and training process for the proposed contrastive objective. 

\begin{figure}[ht]
    \centering
    \begin{subfigure}[b]{0.9\columnwidth}
        \centering
        \resizebox{\columnwidth}{!}{%
        \begin{tabular}{ll}
        \multirow{2}{*}{\textcolor{ForestGreen}{\textbf{Sample 1} (+)}} & \textcolor{Blue}{\textbf{U}: Play }\\
        & \textcolor{Blue}{\textbf{R}: What do you want me to play?} \\
        \\
        \multirow{2}{*}{\textcolor{ForestGreen}{\textbf{Sample 2} (+)}} & \textcolor{Blue}{\textbf{U}: What time is it?}\\
        & \textcolor{Blue}{\textbf{R}: The time is 12:55 pm} \\
        \\
        \multirow{2}{*}{\textcolor{BrickRed}{\textbf{Sample 3} (-)}} & \textcolor{Blue}{\textbf{U}: Play }\\
        & \textcolor{Blue}{\textbf{R}: The time is 12:55 pm} \\
        \\
        \multirow{2}{*}{\textcolor{BrickRed}{\textbf{Sample 4} (-)}} & \textcolor{Blue}{\textbf{U}: What time is it?}\\
        & \textcolor{Blue}{\textbf{R}: What do you want me to play?} \\
        \end{tabular}
        }
    \end{subfigure}
    \caption{A toy example demonstrating the generation of unreal samples from a batch of two real samples. 
    Session context is omitted for brevity.
    }
    \label{fig:nce_example}
    \vspace{-0.0in}
\end{figure}

\begin{algorithm}[ht]
   \caption{Contrastive Self-Supervised Training}
   \label{alg:nce}
\begin{algorithmic}
   \STATE {\bfseries Input:} $\mathbb{D}_{unsup}$, $h_{\theta}$ (model w/ contrastive head)
   \REPEAT
       \STATE $X \leftarrow GetBatch(\mathbb{D}_{unsup})$
       \STATE $batchsize \leftarrow length(X)$
       \STATE $y \leftarrow ones(batchsize)$
       \STATE $X_n \leftarrow clone(X)$
       \IF{$rand() < 0.5$}
            \STATE  $shuffle(X_n[`targeted\_utterance`])$
       \ELSE
            \STATE  $shuffle(X_n[`targeted\_response`])$
       \ENDIF
       \STATE $y_n \leftarrow zeros(batchsize)$
       \STATE $p \leftarrow h_{\theta}([X;X_n])$
       \STATE $loss \leftarrow BCE(p, [y;y_n])$
        \STATE Backprop $loss$
       \STATE Update $\theta$
   \UNTIL{$Max Epoch$}
\end{algorithmic}
\end{algorithm}

\subsection{Contrastive Pretraining}
The objective introduced in Section~\ref{sec:nce} is not directly applicable to be used as a user satisfaction model. One approach to leverage the pool of unsupervised data is to pre-train the model on unlabeled data using the self-supervised objective, and then attach a classifier head and finetune the network to distinguish SAT and DSAT samples. In our implementation, we pre-train using the self-supervised objective on $\mathbb{D}_{unsup}$ for $10$ epochs, then train a classifier head on $\mathbb{D}_{sup}$ for another $10$ epochs; adjusting the learning rates for the network body to $\times 0.1$ of the base learning rates (see Section~\ref{sec:sup} for more information on the learning rate setup).

\subsection{Few-Shot Learning}
In the pretraining approach, we solely relied on the loose semantic relationship between the self-supervised and the user satisfaction modeling tasks. However, it is desirable to have a representation that is not only solving the self-supervised task but is also useful for the final objective. In other words, we have a source task ($S$) which we have a large number of training samples and a target task ($T$) with a limited number of samples that is our main interest. The idea is to use information from the target task during the source training such that the trained model is most compatible with the target.

Let us assume we have datasets $\mathbb{D}_S$ and $\mathbb{D}_T$ corresponding to the source ($S$) and target ($T$) tasks as well as inference functions for each task: $f_S(.|\theta,\omega_S)$ and $f_T(.|\theta,\omega_T)$. In this notation, $\theta$ represents shared network parameters (i.e., the body in our architecture) and $\omega$ represents task-specific parameters (i.e., a head in our architecture). Formally, when optimizing for task $S$, we are interested in:
\begin{equation}
    \argmin_{\theta,\omega_S} \mathbb{E}_{\bm{x},y \sim \mathbb{D}_S} [ L_S (f_S (\bm{x}|\theta,\omega_S), y)] \;,
\end{equation}

where $L_S$ is the loss function for the source task. A simple gradient descent step to solve this problem can be written as:
\begin{equation}
\begin{aligned}
    \theta^{t+1} \longleftarrow \theta^{t} - \eta \nabla_{\theta} \mathbb{E}(L_S(f_S (\bm{x}|\theta^t,\omega_S^t), y)) \;,\\
    \omega_S^{t+1} \longleftarrow \omega_S^{t} - \eta \nabla_{\omega_S} \mathbb{E}(L_S(f_S (\bm{x}|\theta^t,\omega_S^t), y)) \;.
\end{aligned}
\end{equation}

However, we are interested in optimization steps that do not increase the loss value for task $T$:
\begin{equation}
\label{eq:not_increase}
\begin{aligned}
    \mathbb{E}_{\bm{x},y \sim \mathbb{D}_T} [ L_{T} (f_T (\bm{x}|\theta^{t+1},\omega_T^{t+1}), y)] \leq \\
    \mathbb{E}_{\bm{x},y \sim \mathbb{D}_T} [ L_{T} (f_T (\bm{x}|\theta^{t},\omega_T^{t}), y)] \;.
\end{aligned}
\end{equation}

Considering \eqref{eq:not_increase} as an optimization constraint can potentially halt the optimization because improvements to the source objective do not directly translate to improvements to the target task. In other words, the constraint above may not be always directly satisfiable using gradient steps in the source domain.

To overcome this issue, instead of using gradient descent, we define the problem as a Randomize Block Coordinate Descent (RBCD)~\citep{nesterov2012efficiency,wright2015coordinate} optimization. At each RBCD iteration, only a subset of model parameters, i.e. a block noted as $\bm{b}$, is sampled from a distribution $\mathcal{B}$ and used for the gradient descent update\footnote{Note that the block selection operation is discrete, either a certain parameter belongs to the block or not, but the distribution $\mathcal{B}$ can be a continuous or discrete probability distribution.}:
\begin{equation}
\begin{aligned}
    \bm{b} \sim \mathcal{B} ~~~~~~~~~~~~~~~~~~~~~~~~~~~~~ \;,\\
    \theta^{t+1}_{\bm{b}} \longleftarrow \theta^{t}_{\bm{b}} - \eta \nabla_{\bm{b}} \mathbb{E}(L_S(f_S (\bm{x}|\theta^t,\omega_S^t), y)) \;.
\end{aligned}
\end{equation}
Note that we only use the RBCD optimization for the network body parameters ($\theta$), while the head parameters ($\omega_S$ and $\omega_T$) are optimized using a regular gradient descent optimization.

In this work, we propose the idea of adjusting the block selection distribution, $\mathcal{B}$, such that parameters having more aligned source and target gradients have more chance of being selected:
\begin{equation}
\label{eq:block_sel}
\begin{aligned}
    \mathcal{B}: Pr(i \in \bm{b}) \propto \langle \nabla_{i,S}L_S, \nabla_{i,T}L_T \rangle \;,
\end{aligned}
\end{equation}
where the inputs to $L_S$ and $L_T$ are omitted for brevity. Intuitively, \eqref{eq:block_sel} is used to discourage parameter updates that are not aligned with the T task which can be viewed as a soft method to enforce the constraint in \eqref{eq:not_increase}. 
Here, there are multiple options to define the granularity of the block selection such as layer-wise, neuron-wise, or element-wise. Based on our initial experiments, we found that defining the block elements to be layer-wise results in the best performance.

Algorithm~\ref{alg:joint} shows an outline of the proposed method. At each iteration within the training loop, we back-propagate the $S$ and $T$ losses and store the gradients of layer parameters. For parameters related to the $S$ head, we follow a simple gradient descent update. For body parameters, we only update the parameters if the inner product of the $S$ and $T$ tasks is positive or at a small random outcome with the probability of $\alpha$. 
To guarantee the convergence of the source task, we allow all parameters to be selected at each step at least with a very small probability of $\alpha$.
In our experiments, we consider $\alpha$ as a hyperparameter taking values in $\{0.001, 0.005, 0.01, 0.05, 0.1\}$. Additional care is required when updating the $T$ head layer parameters as the $\mathbb{D}_T$ is usually much smaller than $\mathbb{D}_S$ and the $T$ head is prone to overfitting. We use a validation set from task $T$ to detect overfitting for the $T$ head and early stop the updates. Note that a hyperparameter $\lambda$ is used to set the frequency of the $T$ head updates after the early stopping. Having less frequent head updates allows the $T$ head to gradually improve and adapt to the changes in the body without getting overfitted. In our experiments, we search for proper $\lambda$ values in $\{0.001, 0.002, 0.005,0.01\}$.

\begin{algorithm}[ht]
   \caption{The Proposed Few-Shot Training}
   \label{alg:joint}
\begin{algorithmic}
   \STATE {\bfseries Input:} $\mathbb{D}_S$, $\mathbb{D}_T$, $f_S$, $f_T$, $\alpha$ (random selection rate), $\lambda$ ($T$ head update rate)
   \REPEAT
       \STATE $(\bm{x}_S,y_S) \sim \mathbb{D}_S$
       \STATE $(\bm{x}_T,y_T) \sim \mathbb{D}_T$
       \STATE \texttt{// compute \& store gradients}
       \STATE $loss_S \leftarrow L_S(f_S(\bm{x}_S, y_S))$
       \STATE Backprop and store $loss_S$
       \STATE $loss_T \leftarrow L_T(f_T(\bm{x}_T, y_T))$
       \STATE Backprop and store $loss_T$
       \STATE \texttt{// Layer-Wise RBCD update}
       \FOR{$\mathcal{P}$ in $LayerParameters$ :}
            \IF{$\mathcal{P} \in \omega_S$ \texttt{// if S head param}}
                \STATE $\mathcal{P} \leftarrow \mathcal{P} - \eta \nabla_{\mathcal{P}} loss_S$
             \ELSIF{$\mathcal{P} \in \theta$ \texttt{// if body param}}
                \STATE $sim \leftarrow \langle \nabla_{\mathcal{P}} loss_S, \nabla_{\mathcal{P}} loss_T \rangle$
                \IF{$sim > 0$ or $rand() < \alpha$}
                    \STATE $\mathcal{P} \leftarrow \mathcal{P} - \eta \nabla_{\mathcal{P}} loss_S$
                \ENDIF
            \ELSE 
            \STATE \texttt{// if T head parameter}
                \IF{$NotEarlyStopped$ or $rand() < \lambda$}
                    \STATE $\mathcal{P} \leftarrow \mathcal{P} - \eta \nabla_{\mathcal{P}} loss_T$
                \ENDIF
            \ENDIF
       \ENDFOR
       \STATE Validate, update $NotEarlyStopped$
    \UNTIL{$Max Epoch$}
\end{algorithmic}
\end{algorithm}

In contrast to other works in the literature which mostly leverage the alignment of concatenated gradients~\cite{lopez2017gradient,luo2020n}, we propose layer-wise similarity measurements providing more granularity and more adaptability. Also, the suggested approach does not require any inner loop optimization process or gradient projection and hence is scalable to large-scale problems. The only computational and memory overhead is to store the model gradients with respect to each task and to compute inner products between the layer parameters.

The method explained in this section is general to few-shot transfer learning and joint training settings where a large source dataset is being used to achieve representations that are most useful for a final target task. For our use-case, we use the suggested approach considering the source task, $S$, as the self-supervised contrastive objective and the target task, $T$, as the user satisfaction prediction task. In our experiments, after the joint training process, we reinitialize the $T$ head and finetune the network for the $T$ task. We found this approach to be helpful to achieve the best results as the jointly trained $T$ head is often slightly overfitted.

\section{Results}

\subsection{Experimental Setup}
We used PyTorch \citep{paszke2017automatic} to train our models. For each case, we continue the training for the maximum number of epochs (10 in our experiments) and select the best model based on the validation performance. We conducted our experiments on a cluster of 48 NVIDIA V100 GPUs (16 GB memory, 8 GPUs per machine). It took between about 6 hours to 27 hours to run individual experiments, depending on the case.

For each experiment, we report Area Under the ROC Curve (AUC-ROC) and Area Under the Precision-Recall Curve (AUC-PR) as the performance measures. The results for the in-domain and out-of-domain held-out test sets are reported separately. Note that there is an imbalance in the frequency of SAT and DSAT labels, and also there is a difference in the label distribution for the in-domain and out-of-domain test sets. To ensure the statistical significance of the results, each experiment is repeated four times using random initializations reporting the mean and standard deviations.

\subsection{Quantitative Results}
\label{sec:res_quant}
\begin{figure*}[h]
    \centering
    \begin{subfigure}[b]{0.98\columnwidth}
        \centering
        \includegraphics[width=\columnwidth]{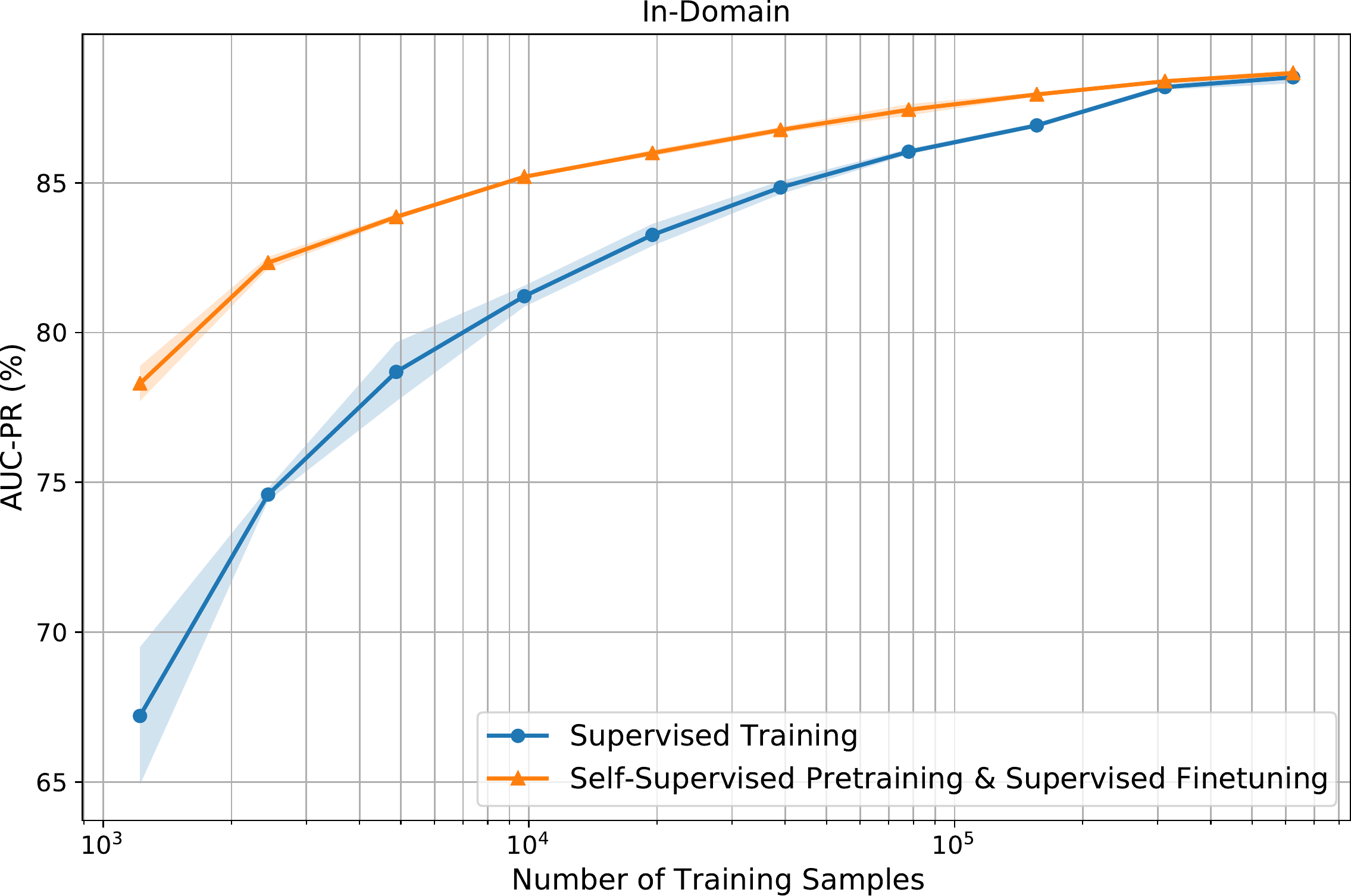}
    \end{subfigure}
    ~
    \begin{subfigure}[b]{0.98\columnwidth}
        \centering
        \includegraphics[width=\columnwidth]{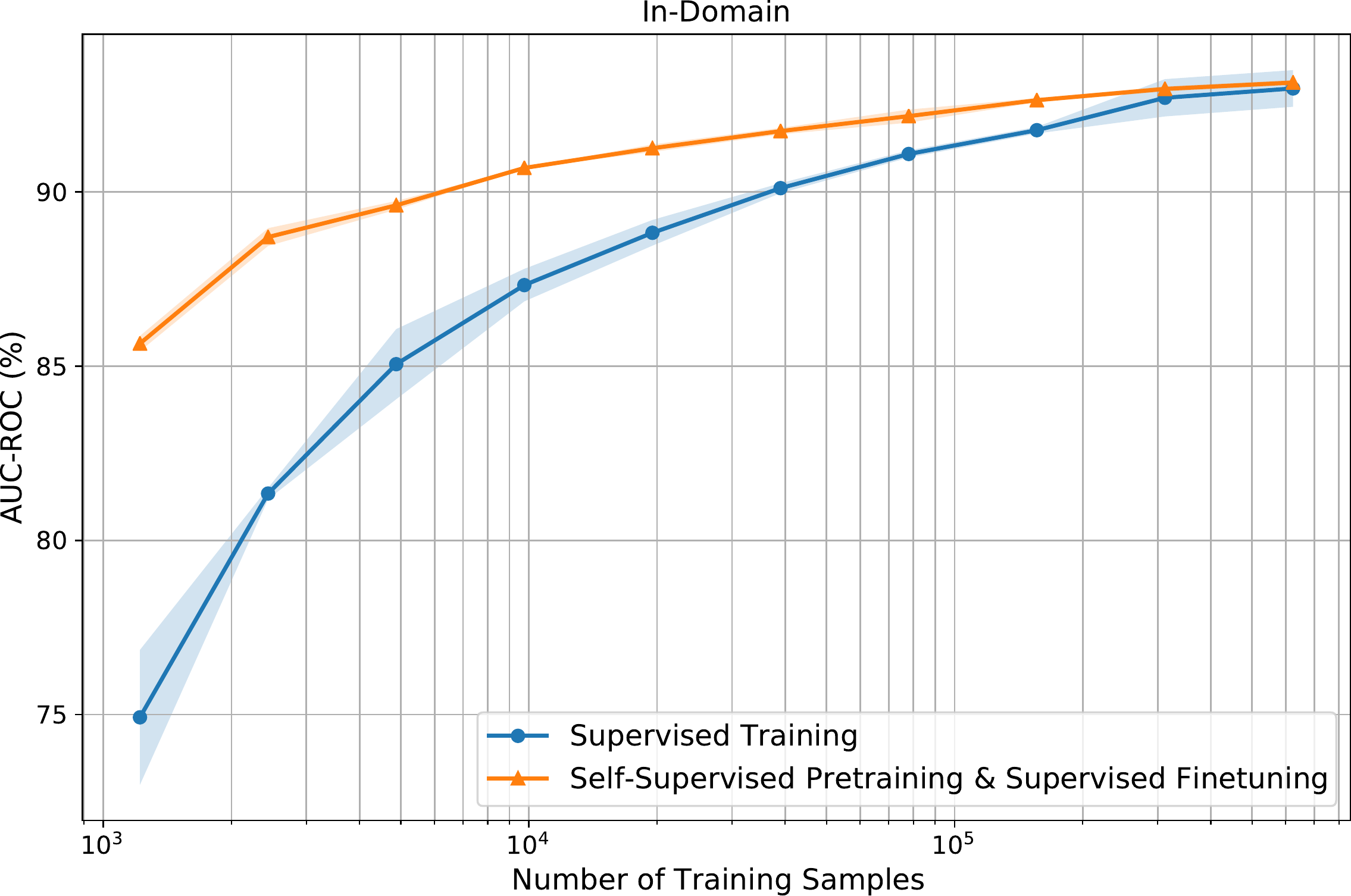}
    \end{subfigure}
    \caption{Comparison of the supervised training baseline and the proposed self-supervised pretraining methods for the in-domain test set using different number of training samples (left: AUC-PR, right: AUC-ROC).}
    \label{fig:sup_vs_nce_indomain}
    \vspace{-0.0in}
\end{figure*}

Figure~\ref{fig:sup_vs_nce_indomain} shows a comparison of the in-domain test results for the supervised training and the self-supervised contrastive pretraining methods. For each case, we report the in-domain test performance using models trained with a different number of annotated training samples. The x-axis is plotted in the log scale. It can be seen that the contrastive self-supervised approach is much more data-efficient compared to the supervised approach as it leverages the pool of unlabeled data. 

\begin{figure*}[h]
    \centering
    \begin{subfigure}[b]{0.98\columnwidth}
        \centering
        \includegraphics[width=\columnwidth]{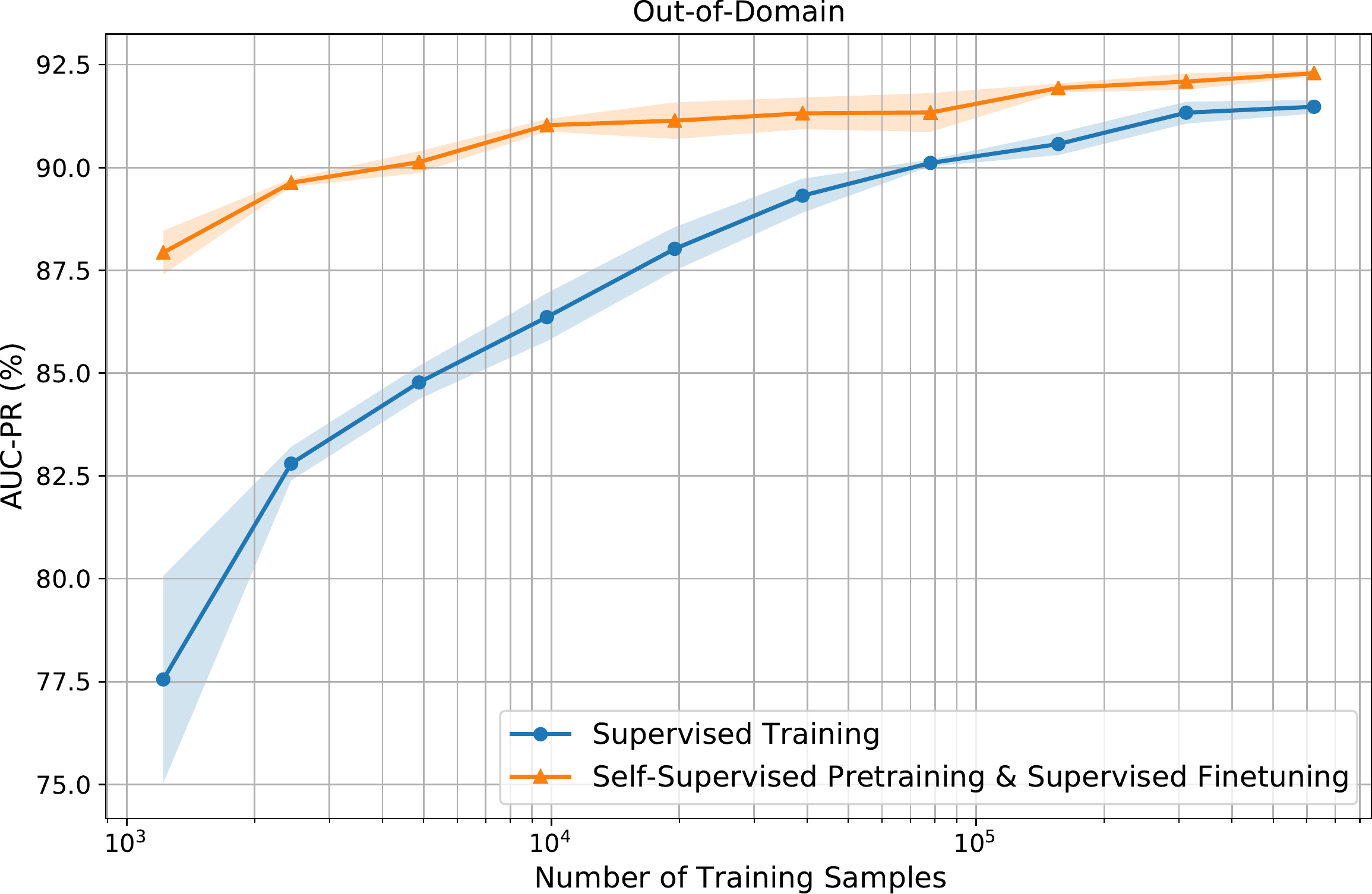}
    \end{subfigure}
    ~
    \begin{subfigure}[b]{0.98\columnwidth}
        \centering
        \includegraphics[width=\columnwidth]{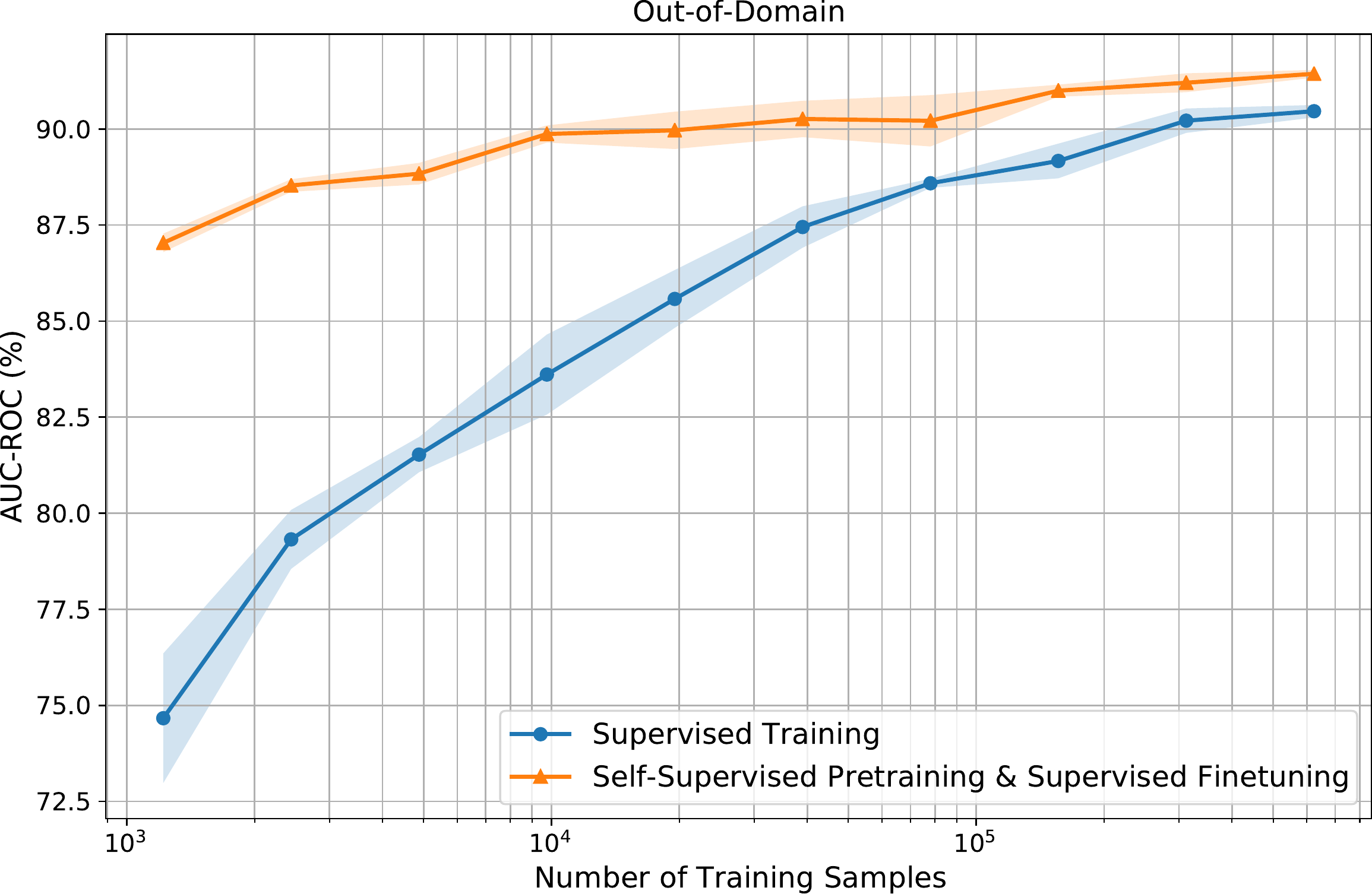}
    \end{subfigure}
    \caption{Comparison of the supervised training baseline and the proposed self-supervised pretraining methods for the out-of-domain test set using different number of training samples (left: AUC-PR, right: AUC-ROC).}
    \label{fig:sup_vs_nce_outofdomain}
    \vspace{-0.0in}
\end{figure*}

Figure~\ref{fig:sup_vs_nce_outofdomain} shows a comparison between the supervised training and the self-supervised pretraining methods on the out-of-domain test set. Similar to the in-domain case, there is a significant gap between the labeled data efficiency of these approaches. However, compared to the in-domain case, using even all training samples, the gap does not appear to close. In other words, for the out-of-domain test set the self-supervised approach is not only more data-efficient but also tends to generalize better. In a real-world conversation system, the out-of-domain generalization can be crucial as many different new skills are being developed and included in the system every day, making the traditional in-domain human annotation less practical due to the required annotation turnaround time.

\begin{figure*}[h]
    \centering
    \begin{subfigure}[b]{0.98\columnwidth}
        \centering
        \includegraphics[width=\columnwidth]{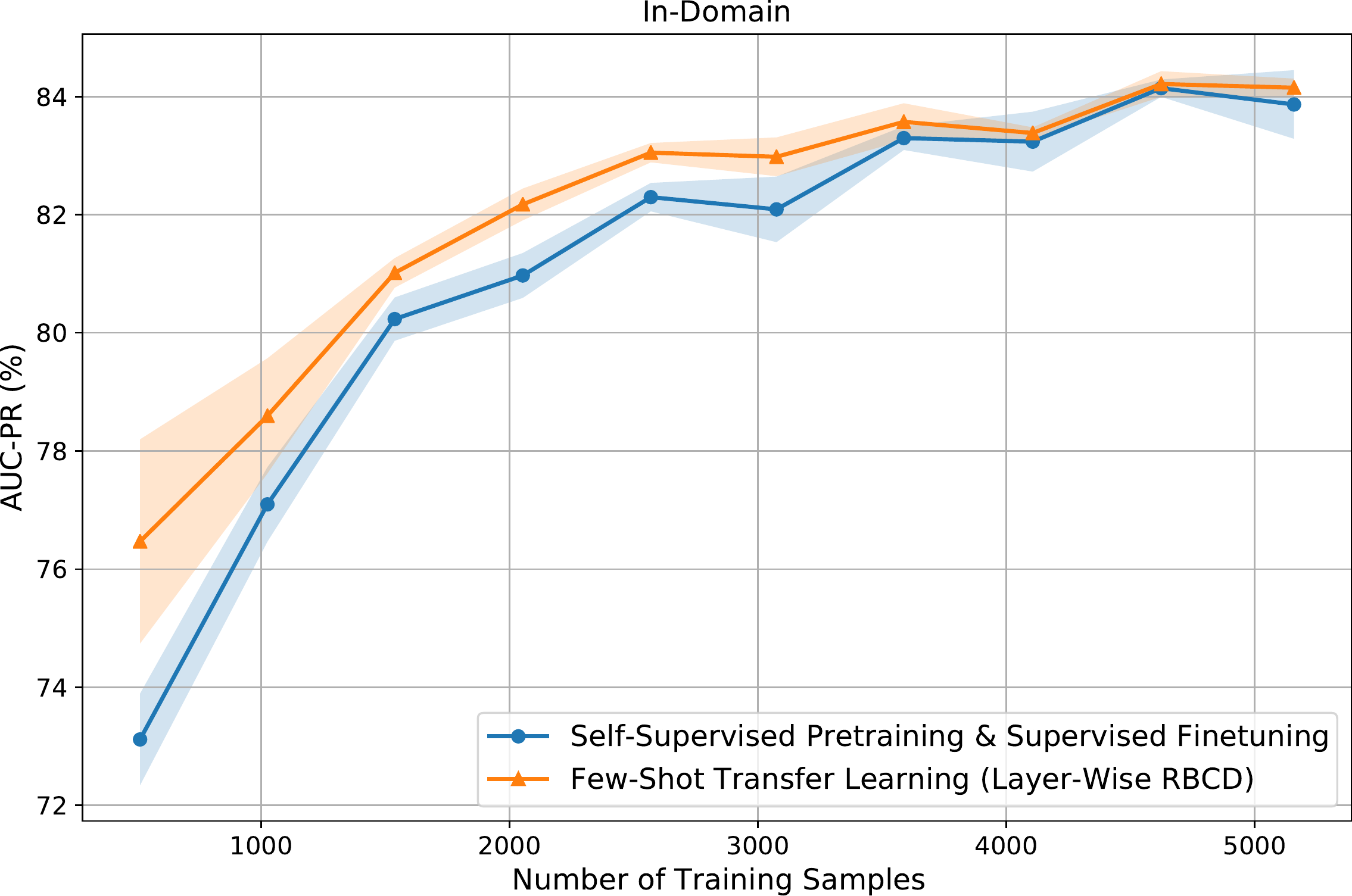}
    \end{subfigure}
    ~
    \begin{subfigure}[b]{0.98\columnwidth}
        \centering
        \includegraphics[width=\columnwidth]{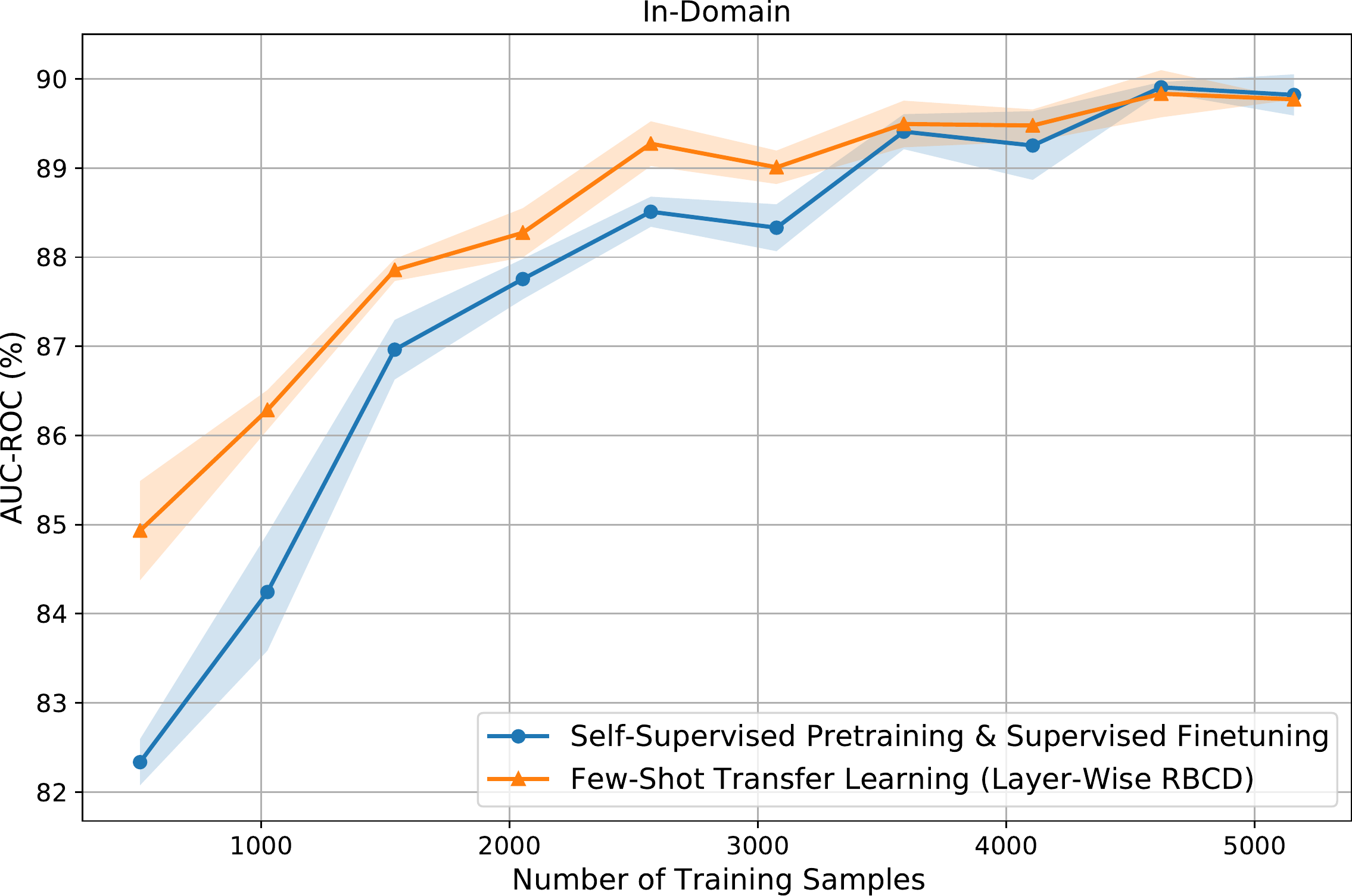}
    \end{subfigure}
    \caption{Comparison of the self-supervised pretraining and the proposed few-shot learning methods for the in-domain test set using different number of training samples (left: AUC-PR, right: AUC-ROC).}
    \label{fig:nce_vs_joint_indomain}
    \vspace{-0.0in}
\end{figure*}

\begin{figure*}[h]
    \centering
    \begin{subfigure}[b]{0.98\columnwidth}
        \centering
        \includegraphics[width=\columnwidth]{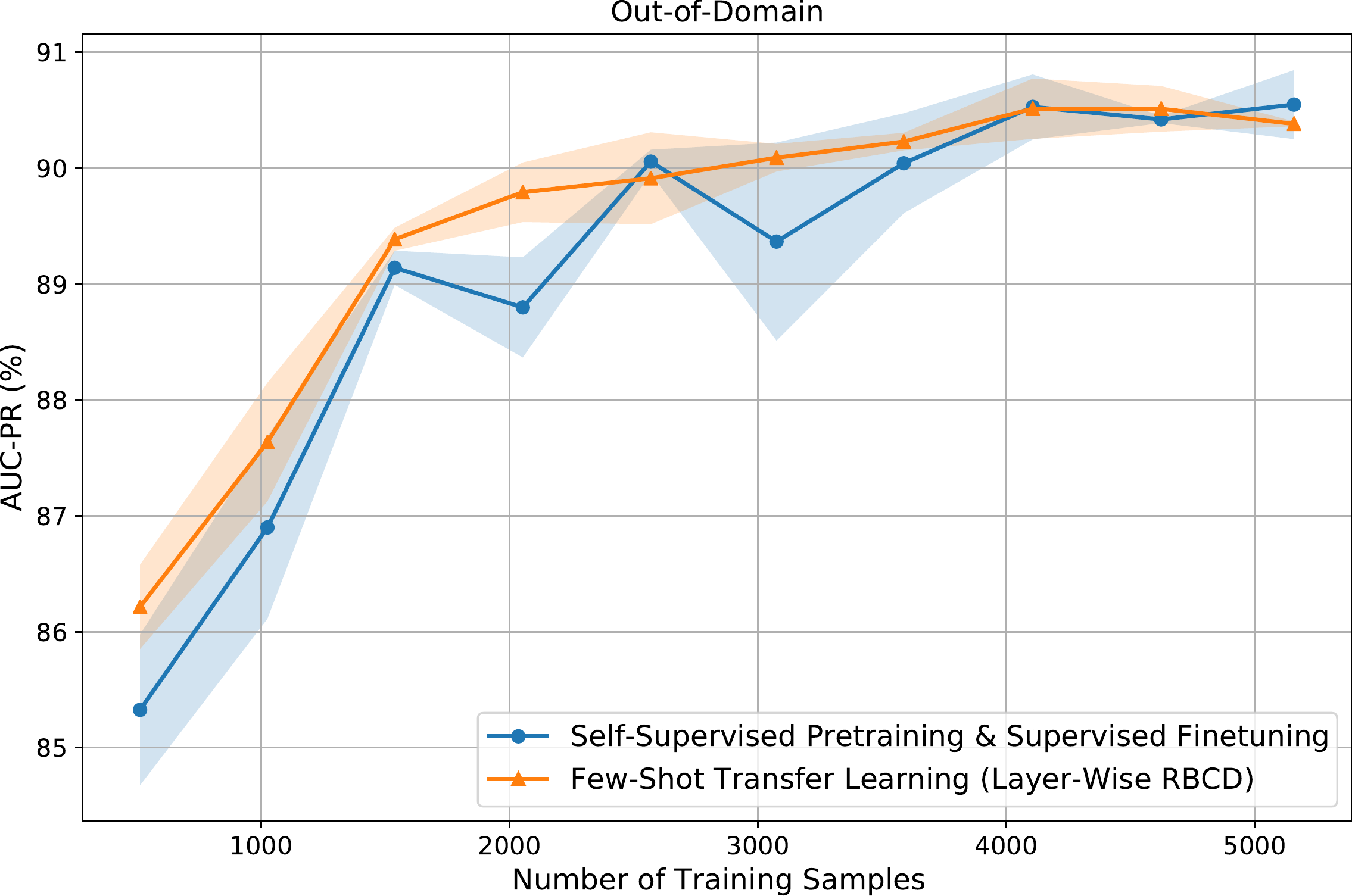}
    \end{subfigure}
    ~
    \begin{subfigure}[b]{0.98\columnwidth}
        \centering
        \includegraphics[width=\columnwidth]{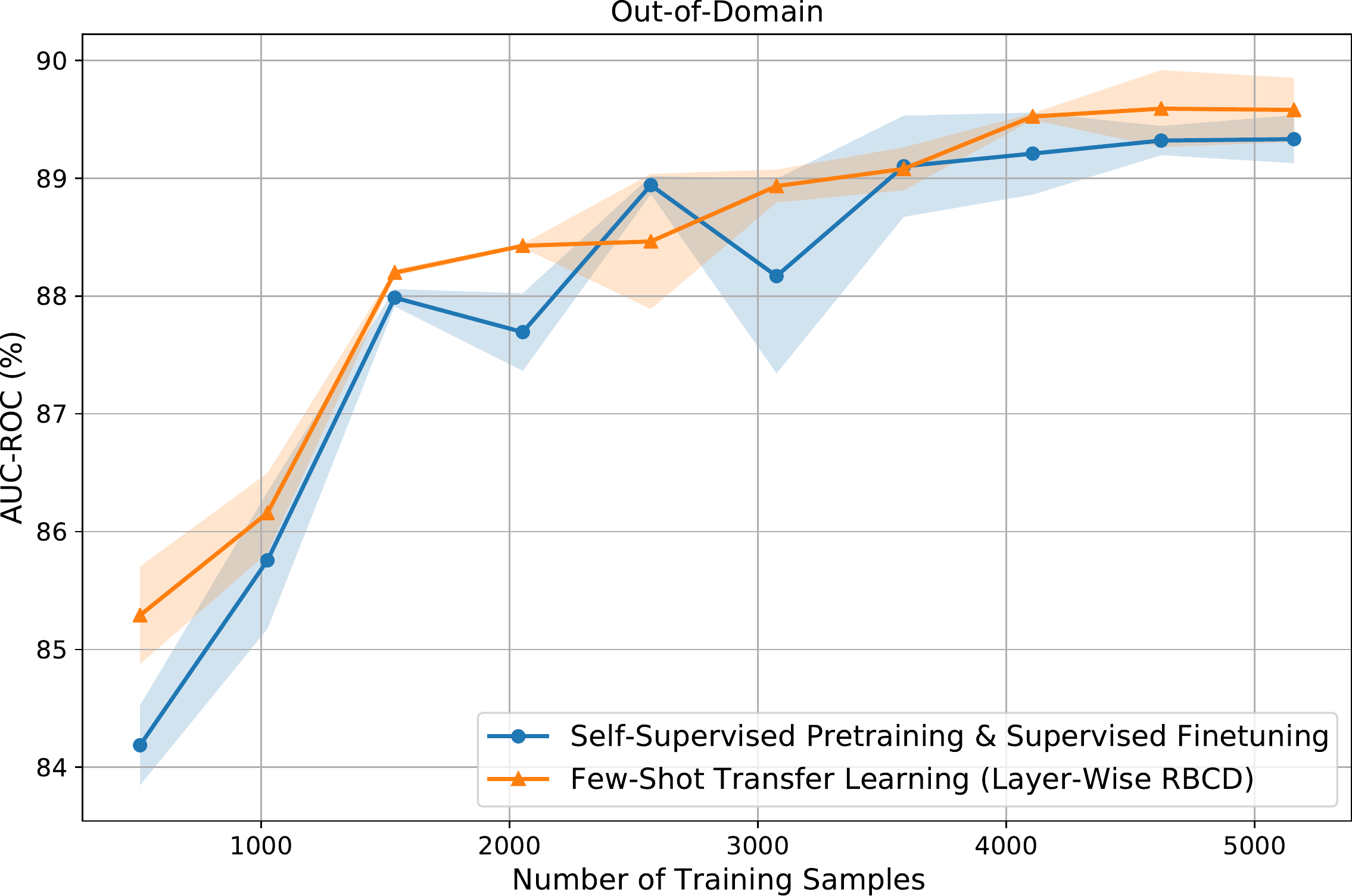}
    \end{subfigure}
    \caption{Comparison of the self-supervised pretraining and the proposed few-shot learning methods for the out-of-domain test set using different number of training samples (left: AUC-PR, right: AUC-ROC).}
    \label{fig:nce_vs_joint_outofdomain}
    \vspace{-0.0in}
\end{figure*}

In Figure~\ref{fig:nce_vs_joint_indomain} and Figure~\ref{fig:nce_vs_joint_outofdomain}, we compare the in-domain and out-of-domain performance of the self-supervised pretraining method with the proposed few-shot learning method. As it can be seen from Figure~\ref{fig:nce_vs_joint_indomain}, the in-domain AUC-PR and AUC-ROC for the few-shot learning are consistently better than the self-supervised pretraining approach. Note that the performance gap closes at about $5000$ samples; perhaps because it is enough training data for fine-tuning and successfully transferring the pre-trained model. The out-of-domain performances as reported in Figure~\ref{fig:nce_vs_joint_outofdomain} show better results for the few-shot approach but the margin of improvement is relatively smaller than the in-domain case.

Note that in the presented results, we focused our comparisons to methods that are scalable and leverage human annotation data for turn-level satisfaction prediction, excluding approaches using human-engineered and skill-specific metrics as well as methods that only consider the quality of conversation from the language perspective.

\subsection{Qualitative Results}
\label{sec:res_qual}

Table~\ref{tab:examples} in Appendix~\ref{sec:Qualitative Results} presents a qualitative comparison of the baseline supervised training and the self-supervised approach suggested in this paper. Here, to highlight the generalization and data-efficiency of each method, we limit the number of annotated samples to $1024$ random samples from the training set of the $\mathbb{D}_{sup}$ dataset. For this table, we provide sample sessions that are chosen with an emphasis on more difficult requests, unclear requests, or requests involving 3p skills. $U$ and $R$ indicate the targeted utterance and response, while $U+x$ and $R+x$ indicate the context utterance and responses appearing $x$ turns after the targeted turn. 

From the provided examples, it can be inferred that the self-supervised approach provides a deeper understanding of the user-agent interaction and is able to generalize better even for infrequent 3p skills. It is consistent with the quantitative results presented in the paper.

\section{Conclusion}
This paper suggested a self-supervised objective to learn user-agent interactions leveraging large amounts of unlabeled data available. 
In addition to the standard fine-tuning approach, this paper presented a novel few-shot transfer learning method based on adjusting the RBCD block selection distribution to favor layer parameters with source and target gradients pointing in similar directions. According to the experiments using real-world data, the proposed approach not only requires significantly less number of annotations, but also generalizes better for unseen out-of-domain skills. 


\FloatBarrier
\bibliography{refs}
\bibliographystyle{./template/naacl2021/acl_natbib}

\clearpage
\appendix

\section{Annotation Protocol}
\label{sec:appendix_annotation}
In the following, we provide a summary of main points considered to produce the annotations used in this paper\footnote{Certain details were omitted to comply with confidentiality requirements.}:
\begin{itemize}
    \item Human annotators were trained to annotate samples i.e., we do not use domain specific metrics or other automated success measures as annotation.
    \item It was made clear to the annotators that the task is turn-level user satisfaction, and not the overall satisfaction over session. Also, instructions were provided on how to handle ASR errors, repeated requests, multiple users in one utterance, and many other special cases.
    \item The annotators were provided the targeted turn as well as a few context turns. This helped them to better understand the actual user intention and judge accordingly.
    \item They were asked to rate the system's response quality in terms of the user satisfaction on the scale of 1 to 5, from terrible to excellent. See Table~\ref{tab:score_examples} for score categories and an example of each category.
    \item To ensure the quality of annotations, each sample was annotated multiple times by different annotators.
\end{itemize}

\begin{table}[h]%
\caption{Examples of the annotation scores.}
\centering
\begin{center}
\resizebox{\columnwidth}{!}{
\begin{tabular}{c|l}
\toprule
\textbf{Score} & \textbf{Example} \\
\hline

1 & U: \textit{Is bacon really bad for your health?} \\
(Terrible) & R: \textit{Sorry, I'm not sure.} \\
\hline

2 & U: \textit{How's the humidity?} \\
(Bad) & R: \textit{Sorry, I don't have daily forecast information.} \\
\hline

3 & U: \textit{Play a song for the baby.} \\
(Okay) & R: \textit{Skill [SKILL NAME] can help you with that.} \\
 & \textit{~~~~Did you want to enable it?} \\
\hline

4 & U: \textit{What's the temperature going to be tomorrow?} \\
(Good) & R: \textit{Tomorrow in London, you can look for cloudy} \\
 & \textit{~~~~weather with a high of 65 and low of 52 degrees.} \\
\hline

5 & U: \textit{Cancel alarm.} \\
(Excellent) & R: \textit{6:35 PM alarm cancelled.} \\
\hline

\bottomrule
\end{tabular}
}
\end{center}
\label{tab:score_examples}
\end{table}%

In our analysis, we consider all samples having a score of 3 or better as SAT, and DSAT otherwise. Also, in our data pipeline, we considered different annotations of the same utterance as different samples. However, care was taken in the data split process to ensure there is no train data contamination in our validation and test sets.

\section{Qualitative Results}
\label{sec:Qualitative Results}
Table~\ref{tab:examples} presents a qualitative comparison of the baseline supervised training and the self-supervised approach suggested in this paper. Here, to highlight the generalization and data-efficiency of each method, we limit the number of annotated samples to $1024$ random samples from the training set of the $\mathbb{D}_{sup}$ dataset. For this table, we provide sample sessions that are chosen with an emphasis on more difficult requests, unclear requests, or requests involving 3p skills. $U$ and $R$ indicate the targeted utterance and response, while $U+x$ and $R+x$ indicate the context utterance and responses appearing $x$ turns after the targeted turn. 

From these examples, it can be inferred that the self-supervised approach provides a deeper understanding of the user-agent interaction and is able to generalize better even for infrequent 3p skills. It is consistent with the quantitative results presented in the paper.

\newcolumntype{L}{>{\arraybackslash}m{5cm}}
\newcolumntype{I}{>{\arraybackslash}m{3cm}}

\begin{table*}[ht]%
\renewcommand{\arraystretch}{1.4}
\caption{Qualitative comparison of predictions for different methods on a set of selected test samples.}
\centering
\begin{minipage}{0.90\textwidth}
\begin{center}
\resizebox{\columnwidth}{!}{
\begin{tabular}{LL|ccc|I}
\toprule
 & & \textbf{Label} & \multicolumn{2}{c}{\textbf{Prediction}} & \textbf{Comments} \\
 \multicolumn{2}{c|}{\textbf{Session}} &  & \textbf{Sup}\footnote{Supervised training baseline} & \textbf{Self}\footnote{Self-supervised contrastive learning method} \\
 \hline

U-1: \textit{what what high school basketball team do you like} & R-1: \textit{I don't have a favorite yet.} &  &  & &  \\
\textbf{U :} \textit{what is your favorite m. l. b. team} & \textbf{R :} \textit{My home team is [NAME], and you've gotta root root root for the home team} & SAT & \textcolor{BrickRed}{DSAT} & \textcolor{ForestGreen}{SAT} & \small{Asking for favorite, agent using advanced language and slang} \\
U+1: \textit{what's your favorite movie} & R+1: \textit{The Empire Strikes Back" is my all-time favorite. It's impressive.} &  &  &  \\
\hline

U-1: \textit{is it going to rain tomorrow} & R-1: \textit{No rain is expected in New York tomorrow.} &  &  & &  \\
\textbf{U :} \textit{is it raining now in new york} & \textbf{R :} \textit{No rain is expected in New York tomorrow} & DSAT & \textcolor{BrickRed}{SAT} & \textcolor{ForestGreen}{DSAT} & \small{Asking for weather, rewording and repeating the request} \\
U+1: \textit{is it raining today} & R+1: \textit{It's raining right now. You can expect about 0.05 inches.} &  &  &  \\
U+2: \textit{set an alarm for five p.m.} & R+1: \textit{Alarm set for 5:00 PM.} &  &  &  \\
\hline

\textbf{U :} \textit{close the garage} & \textbf{R :} \textit{Ok, Here's Garadget - Smart Garage The door 'garage' was already closed} & SAT & \textcolor{BrickRed}{DSAT} & \textcolor{ForestGreen}{SAT} & \small{Device interaction, using a less common 3p skill} \\
\hline

\textbf{U :} \textit{teach me} & \textbf{R :} \textit{Me's definition is first-person singular pronoun, indicating the speaker, used as the object of a verb or preposition} & DSAT & \textcolor{BrickRed}{SAT} & \textcolor{ForestGreen}{DSAT} & \small{Knowledge question, vague request agent fails to understand} \\
U+1: \textit{pause} & R+1: \textit{[EMPTY]} &  &  &  \\
\hline

U-2: \textit{what is in the middle of the milky way galaxy} & R-2: \textit{A storm of exploding stars circling a black hole.} &  &  &  \\
U-1: \textit{name all the countries in the world} & R-1: \textit{There are 193 UN-recognized countries, including China, India, the United States, and Indonesia.} &  &  &  \\
\textbf{U :} \textit{read us a really long list} & \textbf{R :} \textit{Resuming your most recently purchased book, Everything I Know About Lean I Learned in First Grade} & DSAT & \textcolor{BrickRed}{SAT} & \textcolor{ForestGreen}{DSAT} & \small{Knowledge question, agent fails to understand user's intention} \\
U+1: \textit{cancel} & R+1: \textit{[EMPTY]} &  &  &  \\
U+2: \textit{name all the countries in europe} & R+2: \textit{I know about fifty countries in Europe. They are the United Kingdom, Turkey, [48 MORE NAMES]} &  &  &  \\
\hline

U-2: \textit{call} & R-2: \textit{Which Phone Number, Contact or Device do you want to call?} &  &  &  \\
U-1: \textit{p.m} & R-1: \textit{I heard you ask for p.m., but I can't find that name in your list of contacts or devices.} &  &  &  \\
\textbf{U :} \textit{call liam} & \textbf{R :} \textit{I can't call Liam's phone number. I can call contacts who have signed up for [AGENT] calling, and I can call phone numbers in the United States, Canada, and Mexico} & SAT & \textcolor{BrickRed}{DSAT} & \textcolor{ForestGreen}{SAT} & \small{Call request, failed due to system limitation but user understands} \\
U+1: \textit{thank you [AGENT]} & R+1: \textit{You bet . And good evening.} &  &  &  \\

\bottomrule
\end{tabular}
}
\end{center}
\end{minipage}
\label{tab:examples}
\end{table*}%

\end{document}

%% file: template/math_commands.tex
\usepackage{amsmath,amsfonts,bm}

\def\1{\bm{1}}

\DeclareMathAlphabet{\mathsfit}{\encodingdefault}{\sfdefault}{m}{sl}
\SetMathAlphabet{\mathsfit}{bold}{\encodingdefault}{\sfdefault}{bx}{n}

\DeclareMathOperator*{\argmin}{arg\,min}

%% file: main.bbl
\begin{thebibliography}{31}
\expandafter\ifx\csname natexlab\endcsname\relax\def\natexlab#1{#1}\fi

\bibitem[{Bodigutla et~al.(2019)Bodigutla, Polymenakos, and
  Matsoukas}]{bodigutla2019multi}
Praveen~Kumar Bodigutla, Lazaros Polymenakos, and Spyros Matsoukas. 2019.
\newblock Multi-domain conversation quality evaluation via user satisfaction
  estimation.
\newblock \emph{arXiv preprint arXiv:1911.08567}.

\bibitem[{Chung et~al.(2014)Chung, Gulcehre, Cho, and
  Bengio}]{chung2014empirical}
Junyoung Chung, Caglar Gulcehre, KyungHyun Cho, and Yoshua Bengio. 2014.
\newblock Empirical evaluation of gated recurrent neural networks on sequence
  modeling.
\newblock \emph{arXiv preprint arXiv:1412.3555}.

\bibitem[{Devlin et~al.(2018)Devlin, Chang, Lee, and
  Toutanova}]{devlin2018bert}
Jacob Devlin, Ming-Wei Chang, Kenton Lee, and Kristina Toutanova. 2018.
\newblock Bert: Pre-training of deep bidirectional transformers for language
  understanding.
\newblock \emph{arXiv preprint arXiv:1810.04805}.

\bibitem[{Devon et~al.(2020)}]{devon2020representation}
R~Devon et~al. 2020.
\newblock Representation learning with video deep infomax.
\newblock \emph{arXiv preprint arXiv:2007.13278}.

\bibitem[{Fang and Xie(2020)}]{fang2020cert}
Hongchao Fang and Pengtao Xie. 2020.
\newblock Cert: Contrastive self-supervised learning for language
  understanding.
\newblock \emph{arXiv preprint arXiv:2005.12766}.

\bibitem[{Fox et~al.(2005)Fox, Karnawat, Mydland, Dumais, and
  White}]{fox2005evaluating}
Steve Fox, Kuldeep Karnawat, Mark Mydland, Susan Dumais, and Thomas White.
  2005.
\newblock Evaluating implicit measures to improve web search.
\newblock \emph{ACM Transactions on Information Systems (TOIS)},
  23(2):147--168.

\bibitem[{Gutmann and Hyv{\"a}rinen(2010)}]{gutmann2010noise}
Michael Gutmann and Aapo Hyv{\"a}rinen. 2010.
\newblock Noise-contrastive estimation: A new estimation principle for
  unnormalized statistical models.
\newblock In \emph{Proceedings of the Thirteenth International Conference on
  Artificial Intelligence and Statistics}, pages 297--304.

\bibitem[{Hassan(2012)}]{hassan2012semi}
Ahmed Hassan. 2012.
\newblock A semi-supervised approach to modeling web search satisfaction.
\newblock In \emph{Proceedings of the 35th international ACM SIGIR conference
  on Research and development in information retrieval}, pages 275--284.

\bibitem[{Hjelm et~al.(2018)Hjelm, Fedorov, Lavoie-Marchildon, Grewal, Bachman,
  Trischler, and Bengio}]{hjelm2018learning}
R~Devon Hjelm, Alex Fedorov, Samuel Lavoie-Marchildon, Karan Grewal, Phil
  Bachman, Adam Trischler, and Yoshua Bengio. 2018.
\newblock Learning deep representations by mutual information estimation and
  maximization.
\newblock In \emph{International Conference on Learning Representations}.

\bibitem[{J{\"a}rvelin et~al.(2008)J{\"a}rvelin, Price, Delcambre, and
  Nielsen}]{jarvelin2008discounted}
Kalervo J{\"a}rvelin, Susan~L Price, Lois~ML Delcambre, and Marianne~Lykke
  Nielsen. 2008.
\newblock Discounted cumulated gain based evaluation of multiple-query ir
  sessions.
\newblock In \emph{European Conference on Information Retrieval}, pages 4--15.
  Springer.

\bibitem[{Jiang et~al.(2015)Jiang, Hassan~Awadallah, Jones, Ozertem, Zitouni,
  Gurunath~Kulkarni, and Khan}]{jiang2015automatic}
Jiepu Jiang, Ahmed Hassan~Awadallah, Rosie Jones, Umut Ozertem, Imed Zitouni,
  Ranjitha Gurunath~Kulkarni, and Omar~Zia Khan. 2015.
\newblock Automatic online evaluation of intelligent assistants.
\newblock In \emph{Proceedings of the 24th International Conference on World
  Wide Web}, pages 506--516.

\bibitem[{Kingma and Ba(2014)}]{kingma2014adam}
Diederik~P Kingma and Jimmy Ba. 2014.
\newblock Adam: A method for stochastic optimization.
\newblock \emph{arXiv preprint arXiv:1412.6980}.

\bibitem[{Li et~al.(2020)Li, Xu, Yongkang, Zhao, and
  Kankanhalli}]{li2020gradmix}
Junnan Li, Ziwei Xu, Wong Yongkang, Qi~Zhao, and Mohan Kankanhalli. 2020.
\newblock Gradmix: Multi-source transfer across domains and tasks.
\newblock In \emph{The IEEE Winter Conference on Applications of Computer
  Vision}, pages 3019--3027.

\bibitem[{Lin(2004)}]{lin2004rouge}
Chin-Yew Lin. 2004.
\newblock Rouge: A package for automatic evaluation of summaries.
\newblock In \emph{Text summarization branches out}, pages 74--81.

\bibitem[{Liu et~al.(2016)Liu, Lowe, Serban, Noseworthy, Charlin, and
  Pineau}]{liu2016not}
Chia-Wei Liu, Ryan Lowe, Iulian~V Serban, Michael Noseworthy, Laurent Charlin,
  and Joelle Pineau. 2016.
\newblock How not to evaluate your dialogue system: An empirical study of
  unsupervised evaluation metrics for dialogue response generation.
\newblock \emph{arXiv preprint arXiv:1603.08023}.

\bibitem[{Liu et~al.(2020)Liu, Zhang, Hou, Wang, Mian, Zhang, and
  Tang}]{liu2020self}
Xiao Liu, Fanjin Zhang, Zhenyu Hou, Zhaoyu Wang, Li~Mian, Jing Zhang, and Jie
  Tang. 2020.
\newblock Self-supervised learning: Generative or contrastive.
\newblock \emph{arXiv}, pages arXiv--2006.

\bibitem[{Liu et~al.(2019)Liu, Ott, Goyal, Du, Joshi, Chen, Levy, Lewis,
  Zettlemoyer, and Stoyanov}]{liu2019roberta}
Yinhan Liu, Myle Ott, Naman Goyal, Jingfei Du, Mandar Joshi, Danqi Chen, Omer
  Levy, Mike Lewis, Luke Zettlemoyer, and Veselin Stoyanov. 2019.
\newblock Roberta: A robustly optimized bert pretraining approach.
\newblock \emph{arXiv preprint arXiv:1907.11692}.

\bibitem[{Lopez-Paz and Ranzato(2017)}]{lopez2017gradient}
David Lopez-Paz and Marc'Aurelio Ranzato. 2017.
\newblock Gradient episodic memory for continual learning.
\newblock In \emph{Advances in neural information processing systems}, pages
  6467--6476.

\bibitem[{Luo et~al.(2020)Luo, Wong, Kankanhalli, and Zhao}]{luo2020n}
Yan Luo, Yongkang Wong, Mohan~S Kankanhalli, and Qi~Zhao. 2020.
\newblock $ n $-reference transfer learning for saliency prediction.
\newblock \emph{arXiv preprint arXiv:2007.05104}.

\bibitem[{Nesterov(2012)}]{nesterov2012efficiency}
Yu~Nesterov. 2012.
\newblock Efficiency of coordinate descent methods on huge-scale optimization
  problems.
\newblock \emph{SIAM Journal on Optimization}, 22(2):341--362.

\bibitem[{Novikova et~al.(2017)Novikova, Du{\v{s}}ek, Curry, and
  Rieser}]{novikova2017we}
Jekaterina Novikova, Ond{\v{r}}ej Du{\v{s}}ek, Amanda~Cercas Curry, and Verena
  Rieser. 2017.
\newblock Why we need new evaluation metrics for nlg.
\newblock \emph{arXiv preprint arXiv:1707.06875}.

\bibitem[{Oord et~al.(2018)Oord, Li, and Vinyals}]{oord2018representation}
Aaron van~den Oord, Yazhe Li, and Oriol Vinyals. 2018.
\newblock Representation learning with contrastive predictive coding.
\newblock \emph{arXiv preprint arXiv:1807.03748}.

\bibitem[{Papineni et~al.(2002)Papineni, Roukos, Ward, and
  Zhu}]{papineni2002bleu}
Kishore Papineni, Salim Roukos, Todd Ward, and Wei-Jing Zhu. 2002.
\newblock Bleu: a method for automatic evaluation of machine translation.
\newblock In \emph{Proceedings of the 40th annual meeting of the Association
  for Computational Linguistics}, pages 311--318.

\bibitem[{Park et~al.(2020)Park, Yuan, Kim, Zhang, Spyros, Kim, Sarikaya, Guo,
  Ling, Quinn et~al.}]{park2020large}
Dookun Park, Hao Yuan, Dongmin Kim, Yinglei Zhang, Matsoukas Spyros, Young-Bum
  Kim, Ruhi Sarikaya, Edward Guo, Yuan Ling, Kevin Quinn, et~al. 2020.
\newblock Large-scale hybrid approach for predicting user satisfaction with
  conversational agents.
\newblock \emph{arXiv preprint arXiv:2006.07113}.

\bibitem[{Paszke et~al.(2017)Paszke, Gross, Chintala, Chanan, Yang, DeVito,
  Lin, Desmaison, Antiga, and Lerer}]{paszke2017automatic}
Adam Paszke, Sam Gross, Soumith Chintala, Gregory Chanan, Edward Yang, Zachary
  DeVito, Zeming Lin, Alban Desmaison, Luca Antiga, and Adam Lerer. 2017.
\newblock Automatic differentiation in pytorch.

\bibitem[{Pragst et~al.(2017)Pragst, Ultes, and Minker}]{pragst2017recurrent}
Louisa Pragst, Stefan Ultes, and Wolfgang Minker. 2017.
\newblock Recurrent neural network interaction quality estimation.
\newblock In \emph{Dialogues with Social Robots}, pages 381--393. Springer.

\bibitem[{Rach et~al.(2017)Rach, Minker, and Ultes}]{rach2017interaction}
Niklas Rach, Wolfgang Minker, and Stefan Ultes. 2017.
\newblock Interaction quality estimation using long short-term memories.
\newblock In \emph{Proceedings of the 18th Annual SIGdial Meeting on Discourse
  and Dialogue}, pages 164--169.

\bibitem[{Ram et~al.(2018)Ram, Prasad, Khatri, Venkatesh, Gabriel, Liu, Nunn,
  Hedayatnia, Cheng, Nagar et~al.}]{ram2018conversational}
Ashwin Ram, Rohit Prasad, Chandra Khatri, Anu Venkatesh, Raefer Gabriel, Qing
  Liu, Jeff Nunn, Behnam Hedayatnia, Ming Cheng, Ashish Nagar, et~al. 2018.
\newblock Conversational ai: The science behind the alexa prize.
\newblock \emph{arXiv preprint arXiv:1801.03604}.

\bibitem[{Trinh et~al.(2019)Trinh, Luong, and Le}]{trinh2019selfie}
Trieu~H Trinh, Minh-Thang Luong, and Quoc~V Le. 2019.
\newblock Selfie: Self-supervised pretraining for image embedding.
\newblock \emph{arXiv preprint arXiv:1906.02940}.

\bibitem[{Wright(2015)}]{wright2015coordinate}
Stephen~J Wright. 2015.
\newblock Coordinate descent algorithms.
\newblock \emph{Mathematical Programming}, 151(1):3--34.

\bibitem[{Yao et~al.(2020)Yao, Yi, Cheng, Yu, Menon, Hong, Chi, Tjoa, Ettinger
  et~al.}]{yao2020self}
Tiansheng Yao, Xinyang Yi, Derek~Zhiyuan Cheng, Felix Yu, Aditya Menon, Lichan
  Hong, Ed~H Chi, Steve Tjoa, Evan Ettinger, et~al. 2020.
\newblock Self-supervised learning for deep models in recommendations.
\newblock \emph{arXiv preprint arXiv:2007.12865}.

\end{thebibliography}
